\documentclass[runningheads]{llncs}

\usepackage{eccv/eccv}

\usepackage[dvipsnames]{xcolor}

\usepackage{graphicx} 
\usepackage{color}
\usepackage{tabularray}
\usepackage{multirow}
\usepackage{algpseudocode}
\usepackage{array}
\usepackage{tabularx}
\usepackage{algorithm}
\usepackage{stfloats}
\usepackage{wrapfig}
\usepackage{pifont}
\definecolor{stratOne}{HTML}{895259}
\definecolor{stratTwo}{HTML}{204d4f}
\newcommand{\HA}[1]{{\color{blue}{\bf HA: }#1}} 
\renewcommand{\HA}[1]{{}}

\usepackage{eccv/eccvabbrv}

\usepackage{graphicx}
\usepackage{booktabs}

\usepackage[accsupp]{axessibility}  

\usepackage[pagebackref,breaklinks,colorlinks,citecolor=eccvblue]{hyperref}

\usepackage{orcidlink}

\begin{document}

\title{Exploring Missing Modality in Multimodal Egocentric Datasets}


\author{Merey Ramazanova\inst{1} \and
Alejandro Pardo\inst{1} \and
Humam Alwassel\inst{2} \and
Bernard Ghanem\inst{1}
}

\authorrunning{M.~Ramazanova et al.}

\institute{King Abdullah University of Science and Technology \\
\email{\{merey.ramazanova, alejandro.pardo, bernard.ghanem\}@kaust.edu.sa} \and
Intelmatix
}

\maketitle

\begin{abstract}
    Multimodal video understanding is crucial for analyzing egocentric videos, where integrating multiple sensory signals significantly enhances action recognition and moment localization. However, practical applications often grapple with incomplete modalities due to factors like privacy concerns, efficiency demands, or hardware malfunctions. Addressing this, our study delves into the impact of missing modalities on egocentric action recognition, particularly within transformer-based models. We introduce a novel concept—Missing Modality Token (MMT)—to maintain performance even when modalities are absent, a strategy that proves effective in the Ego4D, Epic-Kitchens, and Epic-Sounds datasets. Our method mitigates the performance loss, reducing it from its original $\sim30\%$ drop to only $\sim10\%$ when half of the test set is modal-incomplete. Through extensive experimentation, we demonstrate the adaptability of MMT to different training scenarios and its superiority in handling missing modalities compared to current methods. Our research contributes a comprehensive analysis and an innovative approach, opening avenues for more resilient multimodal systems in real-world settings.
\end{abstract}

\section{Introduction}



Multimodal video understanding has been the de facto approach for analyzing egocentric videos. Recent works have shown that the complimentary multisensory signals in egocentric videos are superior for understanding actions~\cite{nagrani2021attention,kazakos2019epic,kazakos2021MTCN,kazakos2021slow,lin2022egocentric} and localizing moments~\cite{ramazanova2023owl,barrios2023localizing,pramanick2023egovlpv2, Hanaudiovisual}. However, multimodal systems need to be practical for real-world applications that could suffer from the incompleteness of modality inputs due to privacy, efficiency, or simply device failures~\cite{lee2023multimodal}. For example, when predicting in real-time using a wearable device, parts of the recordings might be scrapped to preserve the privacy of the bystanders/camera wearer~\cite{grauman2022ego4d, gong2023mmg}. Furthermore, using all sensors could be expensive for a wearable device, opting for cheaper modalities such as audio or IMU~\cite{grauman2023ego}. Thus, studying the impact of missing modalities is crucial for realistic performance expectations.

\begin{figure}[t!]
\begin{center}
\includegraphics[width=\linewidth]{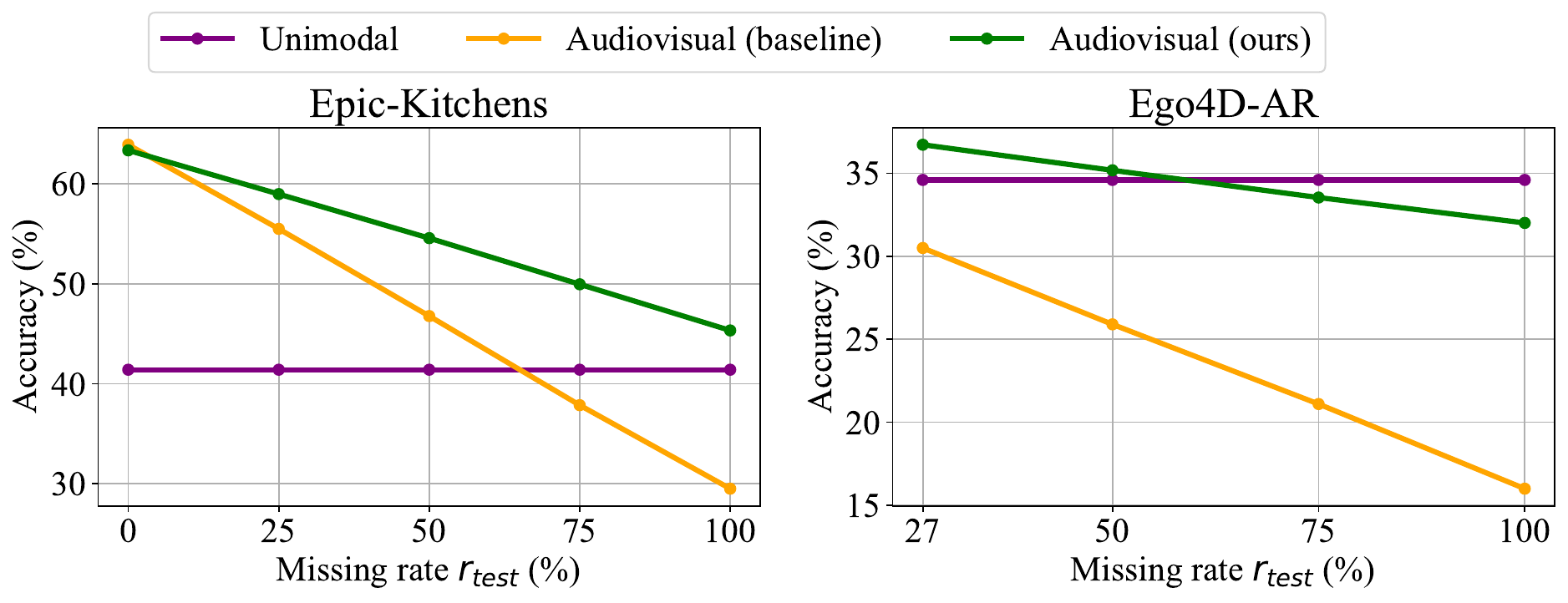}
\end{center}
\small
\caption{Most commonly, we train the multimodal models on modal-complete data. These models (\textcolor{orange}{orange}) fail when encountering modal-incomplete data at test time. Our proposed adaptation to the missing modality (\textcolor{green}{green}) significantly improves the performance across datasets. When all test inputs are modal-incomplete ($r_{test} = 100\%$), we surpass unimodal performance (\textcolor{purple}{purple})  by 5 points in Epic-Kitchens, and double the baseline performance in Ego4D-AR.}
\vspace{-10pt}
\label{figure:intro}
\end{figure}



Still, the current effort to study the impact of missing modalities in egocentric datasets remains rather limited. Most methods presume all modal inputs to be intact during training and inference. Recent works have studied the effect of missing modalities for different tasks varying from recommendation systems to emotion recognition~\cite{neverova2015moddrop,tsai2018learning,ma2022multimodal,wang2018lrmm,lee2019audio,parthasarathy2020training}. Notably, the majority of research concerning missing modalities has primarily addressed the issue during testing~\cite{tsai2018learning, parthasarathy2020training, zhao2021missing,wang2018lrmm,neverova2015moddrop,colombo-etal-2021-improving}, while just a handful studied it across both training and testing phases~\cite{lee2019audio,lee2023multimodal,ma2021smil}. Similar to our setting, Lee~\etal~\cite{lee2023multimodal} propose a strategy to learn prompts for pre-trained backbones to deal with missing modalities. However, they analyze their method for image and text datasets only; we implement our version for action recognition and use it as a baseline in Sec.~\ref{sec:experiments}. More recently, Gong \etal~\cite{gong2023mmg} proposed a benchmark for multimodal generalization, focusing on few-shot learning recognition while considering missing modalities. Though the latter work proposes an interesting benchmark that includes a zero-shot and few-shot setup, no works have diagnosed how recent transformer-based approaches perform when modalities are missing for the action recognition setting.

In this work, we study the problem of missing modalities in egocentric action recognition. First, we investigate how current transformer-based models are affected by incomplete modalities at test time. In Fig.~\ref{figure:intro}, we observe how the current state-of-the-art audiovisual recognition model, Multimodal Bottleneck Transformer (MBT)~\cite{nagrani2021attention}, trained on modal-complete inputs, suffers from a critical degradation in performance when the missing modality rate increases. The advantage of the multimodal backbone (orange) is lost when the missing modality rate in the test set exceeds  $\sim 27\%$ (Ego4D-AR) and $\sim 70\%$ (Epic-Kitchens). At this point, the unimodal model (purple) becomes a better alternative. 
To address this problem, we propose learning the missing modality "template" during training to replace missing modalities at test time. We call this template the Missing Modality Token (MMT) and explain how to learn it in Sec.~\ref{subsec:ours}. Fig.~\ref{figure:intro} (Epic-Kitchens) also shows how our approach (green) dramatically improves the test accuracy and stays at least 5 points above the unimodal performance even when the test set is fully modal-incomplete. Furthermore, our methods enable better multimodal performance overall in modal-incomplete Ego4D-AR. 

We verify the effectiveness of our method in 3 egocentric datasets, including  Ego4D~\cite{grauman2022ego4d}, which has a full coverage of RGB video, but only $70\%$ of videos have audio. 
We extensively analyze our proposed training strategies, showing how to train with MMT under different missing modality scenarios. Our experiments show that our simple yet effective approach proposes a strong solution to this problem. 
\textbf{Our contributions} are threefold: \textbf{(1)} We present a thorough study of the challenge of missing modalities in egocentric action recognition. This involves exploring datasets with varying degrees of modal incompleteness and assessing the influence of the fusion layer.
\textbf{(2)} We propose the Missing Modality Token (MMT) as a novel solution to address missing modalities during both the training and testing phases. Additionally, we propose a training strategy, termed \textit{random-replace}, to enhance the efficacy of models utilizing MMT. \textbf{(3)} We extensively evaluate our method and demonstrate its notable improvement over existing baselines. Through our work, we provide valuable insights and lay the groundwork for developing multimodal backbones that exhibit robustness in the face of missing modalities.
\section{Related work}

\noindent\textbf{Addressing missing modality.}
Addressing missing modalities presents a notable challenge, explored through various strategies by researchers from different areas. From medical applications~\cite{azad2203medical} to sentiment analysis~\cite{colombo-etal-2021-improving}, missing modalities are a long-standing problem in multimodal understanding. Some methods~\cite{radevski2023multimodal, garcia2019learning, garcia2018modality} distill the knowledge from a multimodal teacher to an unimodal RGB model. Others are tailored for scenarios where test data is multimodal yet incomplete in terms of modalities. For example, Ma~\etal~\cite{ma2021smil} and Colombo~\etal~\cite{colombo-etal-2021-improving} investigate missing modalities within a Bayesian Meta-learning framework. Meanwhile, Tsai~\etal~\cite{tsai2018learning}, Zhao~\etal~\cite{zhao2021missing}, and Woo~\etal~\cite{woo2023towards} attempt to reconstruct missing inputs. Neverova~\etal~\cite{neverova2015moddrop} focus on multimodal gesture recognition, employing depth, audio, and video streams, and updating network parameters based on different modality combinations. Most of these works rely on modality-specific architectures~\cite{726791, lstm} and/or use complex generative pipelines. Our approach uses a generic multimodal Transformer~\cite{vaswani2017attention}. Furthermore, these methods assume that the training data is fully modal-complete, which is not the case in current large-scale datasets~\cite{grauman2023ego}. Instead, our method applies to modal-complete and modal-incomplete training sets.


Recent studies utilizing transformers, such as the work of Parthasarathy~\etal~\cite{parthasarathy2020training}, explore missing modalities at test time and propose training-time augmentations.  Ma~\etal~\cite{ma2022multimodal} develop strategies for optimal fusion layers and class tokens in the context of missing modalities, focusing on image-text datasets. Our research differs by demonstrating the effectiveness of our method across various fusion layers (Sec.~\ref{subsec:fusion_layer}), which avoids any expensive fusion policy learning. Lee~\etal~\cite{lee2023multimodal} proposes to learn to prompt large multimodal backbones for image and text classification when modalities are missing at train and test time. We adapt their method to our setting and show that ours is more practical and effective for dealing with missing modalities in egocentric videos.
Lastly, Gong~\etal~\cite{gong2023mmg} introduce a benchmark for handling missing modalities within the Ego4D dataset, tailored for few-shot classification \footnote{Code and data are not available}. 
Our work proposes to diagnose and study the problem in a simpler setting to understand the effect of missing modalities in egocentric video understanding.
We want to note that most of related transformer-based methods~\cite{ gong2023mmg, ma2022multimodal, parthasarathy2020training} do not provide the code, which makes it challenging to compare to.

\noindent\textbf{Multimodal egocentric video understanding.}
Egocentric perception faces distinct challenges compared to traditional video understanding benchmarks such as ActivityNet~\cite{caba2015activitynet} and Kinetics~\cite{kay2017kinetics}. The nature of how egocentric datasets are captured means that they usually feature strongly aligned and synchronized audiovisual signals. Key benchmarks in this field, including Epic-Kitchens~\cite{damen2018scaling}, Epic-Sounds~\cite{huh2023epic}, and the more recent and extensive Ego4D~\cite{grauman2022ego4d}, have demonstrated the importance of audiovisual learning for understanding egocentric videos due to the complementary nature of the audio and visual modalities~\cite{ramazanova2023owl,kazakos2021MTCN, kazakos2019epic}. These datasets have facilitated the creation of several audiovisual backbones tailored for video understanding.
Xiao~\etal~\cite{xiao2020audiovisual} introduced a CNN-based dual-stream architecture, utilizing SlowFast networks for the visual component~\cite{feichtenhofer2019slowfast} and a separate stream for audio~\cite{kazakos2021slow}. With the advent and adaptability of transformer architectures, several studies have treated different modalities as input tokens for a multimodal transformer encoder~\cite{li2020unicoder,kim2021vilt,li2019visualbert, gabeur2020multi}. However, self-attention mechanisms can become prohibitively expensive as the number of tokens increases. To address this, Nagrani~\etal~\cite{nagrani2021attention} offered an efficient Multimodal Bottleneck Transformer (MBT) that avoids costly self-attention. We build atop MBT and introduce  Missing Modality Token to make it robust for missing modalities at train and test times. 
\section{Dealing with missing modalities}
This section details the aspects we consider while addressing the missing modality problem. Namely, the scenarios and evaluation (\ref{subsec:task}), the multimodal design and fusion (\ref{subsec:fusion}), the possible naive solutions to the problem (\ref{subsec:trivial_solutions}), and our proposed method (\ref{subsec:ours}).
\subsection{Problem statement, setup, and evaluation}
\label{subsec:task}
Given the training and testing multimodal data samples, let us denote the missing modality rates in each set with $r_{train}$ and $r_{test}$, respectively. These rates are computed by dividing the number of modal-incomplete samples by the total number of samples. Note that we only consider the setting of one modality being missing in the dataset.
Our work discusses the strategies of training a multimodal model under two scenarios: \textbf{modal-incomplete training set} \ie, some samples in the training set are modal-incomplete ($r_{train} \neq 0\%$), or \textbf{modal-complete training set} \ie, all samples in training set all modal-complete ($r_{train} = 0\%$).

To observe the trained model's behavior under different missing modality severity levels, we create several variants of the test set by manually removing the modality information from the samples until $r_{test} = 100\%$. 

Following previous works~\cite{ma2022multimodal, ma2021smil} when experimenting with fully modal-complete datasets ($r_{train} = 0\%$ and $r_{test} = 0\%$), we assume the modality with the best unimodal performance (dominant) to be incomplete at test-time (\eg, audio for Epic-Sounds). Unlike previous works~\cite{ma2021smil, ma2022multimodal, lee2023multimodal}, we also validate our adaptation strategies on a dataset with naturally incomplete modalities in train and test splits. We use two modalities commonly available in the egocentric video datasets: visual (RGB frames) and audio. We evaluate classification accuracy on egocentric action recognition datasets.




\subsection{Efficient and effective multimodal fusion}
\label{subsec:fusion}
We deal with missing modalities while considering the methods proven to be the most effective for multimodal fusion. Previous transformer-based methods addressing missing modalities looked mostly at basic methods, such as early or mid-fusion with cross-modal self-attention, where all tokens are concatenated at the fusion layer. This does not scale well in videos due to the attention mechanism's quadratic complexity (to the input size)~\cite{nagrani2021attention, lee2023multimodal}. Instead, we use the current state-of-the-art audiovisual fusion, MBT~\cite{nagrani2021attention}, which proved to be more efficient and effective. The bottleneck transformer in the MBT design allows the model to distill and propagate the most essential information across modalities where each modality performs self-attention only with a small number of learnable "bottleneck" tokens. Such design is especially useful for information-dense (redundant) modalities like video.

\begin{figure*}[t!]
\begin{center}
\includegraphics[width=\linewidth]{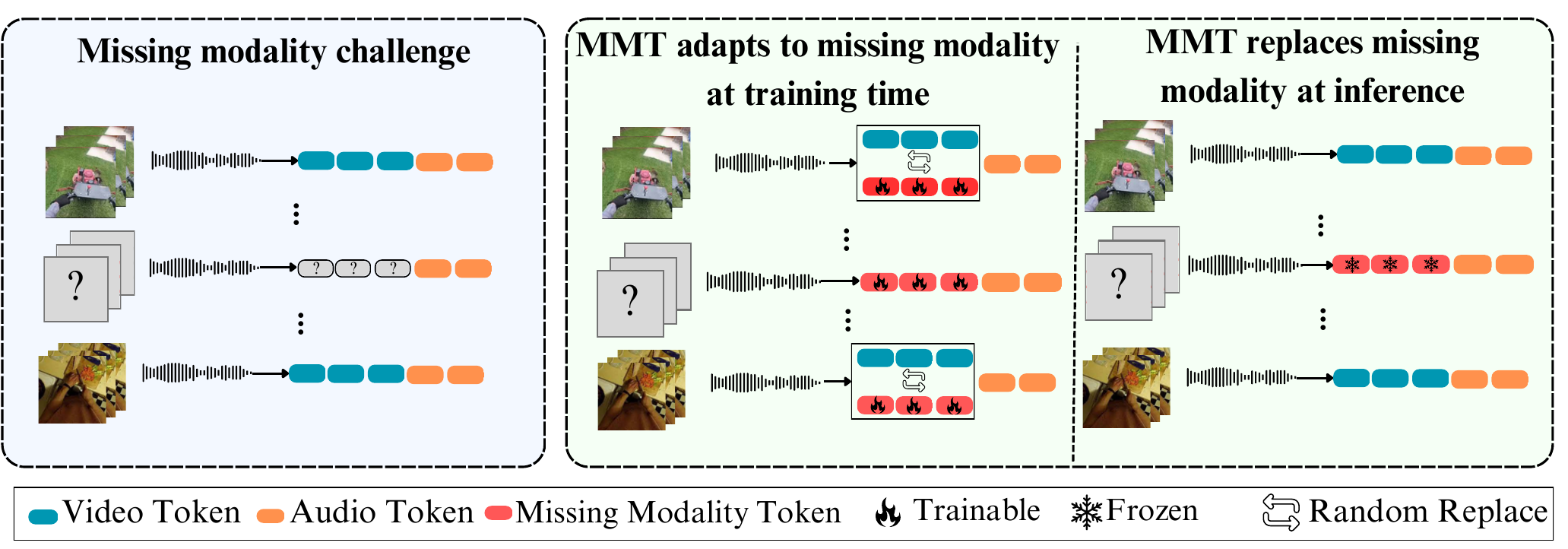}
\end{center}
\small
 \caption{\textbf{Learning and Predicting with Missing Modalities.} \textbf{Left:} Given modal-incomplete data, it is still unclear how to effectively train and predict with a multimodal model (we present some naive baseline methods in Sec.~\ref{subsec:baseline}). \textbf{Right:} To address this issue, we introduce a Missing Modality Token (MMT). During training, MMT learns the representation of missing inputs from modal-incomplete samples and modal-complete samples. For the latter, we use \textit{random-replace} to let the network observe the missing inputs and thus learn better representations (Sec.~\ref{subsec:ours}). At test time, we replace the tokens of missing inputs with MMT to effectively represent them.}
\label{figure:pull}
\end{figure*}

\noindent\textbf{Fusion Layer}. A key aspect of a multimodal fusion approach is the design of the fusion layer $L_f$. $L_f$ is the layer at which the cross-modal interactions happen. We observe the performance of the original bottleneck model with modal-complete audiovisual inputs and train the model with different fusion layers. We show the test accuracy for these models in Figure~\ref{figure:fusion} (marked as \textit{baseline}) for the Epic-Kitchens and Epic-Sounds datasets. We find that the performance does not change significantly with different fusion layers when $r_{test} = 0$. However, the fusion layer does make a difference when inputs are incomplete (\eg, 35\% test accuracy with $L_f = 0$ \vs 42\% with $L_f = 11$ in Epic-Sounds at $r_{test} = 50\%$), \HA{what does $L_f = 0$ and $L_f = 11$ mean? does $L_f$ refer to the number of tokens in the bottleneck? if so, how come it can be 0?} as shown in Section~\ref{subsec:fusion_layer}. Overall, fusing earlier is preferred in Epic-Kitchens, but fusing later gives better results in Epic-Sounds. This outcome is consistent with the observations from the previous work~\cite{ma2022multimodal}: the best fusion strategy is dataset-specific. While this observation might be intuitive, it is impractical when the model is very sensitive to the fusion layer, as searching for the best layer might be computationally expensive. Thus, we analyze the effect of the fusion layer when modalities are missing and show the effect of our approach in Sec.~\ref{subsec:fusion_layer}.

\subsection{Intuitive baselines}\label{subsec:baseline}


\label{subsec:trivial_solutions}
While designing effective solutions for missing modalities is challenging (Fig.~\ref{figure:pull}), one could suggest simple and intuitive ways of adapting to missing inputs at test time. We propose the following \textit{training-free baselines} to deal with missing modality at inference time. 

\noindent\textbf{Passing missing inputs as tensors with zeros.}
We employ a straightforward approach of substituting missing modality inputs with tensors filled with zeroes. This method, widely acknowledged in the literature~\cite{azad2203medical, lee2023multimodal, parthasarathy2020training}, is favored for its simplicity and ease of implementation in practice. Thus, unless stated otherwise, we adopt this method as our primary baseline and refer to it as \textit{baseline}.


\noindent\textbf{Only pass complete inputs.}
As transformers can process sequences of varying lengths, we can selectively omit tokens corresponding to missing signals and exclusively supply the transformer with non-missing modality tokens. While intuitive, this approach becomes less practical when inference involves batch sizes greater than 1 and not all inputs within the batch exhibit modal incompleteness.

 

\subsection{Our approach to dealing with missing modalities}\label{subsec:ours}
We suggest a simple and generic way to deal with missing modalities. Instead of passing tensors filled with zeroes or discarding the tokens of the missing inputs at test time, we propose to learn a "template" for the missing inputs. We introduce a learnable Missing Modality Token (MMT), which is trained to represent the tokens of the missing modality by learning from the tokens of the non-missing modality. 

Figure~\ref{figure:pull} (right) illustrates how MMT represents the tokens of the missing modality at train and test time. When the model encounters a modal-incomplete sample, MMT, a shared parameter, is repeated to match the missing modality token number and passed to the transformer, along with the tokens of non-missing modalities. Note that each token has a different positional embedding added, similar to the mask token in \cite{he2022masked}. With this intuitive way, MMT observes all modal-incomplete samples and leverages the non-missing modality tokens to learn. 

If trained with modal-incomplete samples only, the model never encounters the inputs it tries to mimic. Furthermore, MMT is limited in the number of training samples (\eg in a dataset with $r_{train} = 10\%$, MMT will only encounter $10\%$ of the data). We try to facilitate it and suggest \textit{random-replace} strategy. With \textit{random-replace}, input tokens of modal-complete samples can also be used to train MMT. Namely, for each modal-complete training sample, with probability $p$, the tokens of one modality will be replaced with MMT.  When $p=0$, the MMT learns with naturally missing inputs only. If $p$ is set to a non-zero value, the strategy provides more training samples for learning MMT and lets the network observe the same sample as modal-incomplete and modal-complete; thus encouraging the model to understand the relationships and dependencies of the modalities. However, setting $p$ too high might cause high information loss and hinder the performance, especially if we drop the tokens of a more information-dense modality, as we will show in Sec.~\ref{subsec:random-replace}

Thus, there are 2 ways of providing training samples for MMT: (1) Using samples with naturally missing modalities and (2) Randomly replacing the tokens of complete samples with MMT. Below, we discuss how both ways are used with modal-complete and modal-incomplete training sets.
 
 

\noindent \textbf{Modal-incomplete training set.} 
Recall that $r_{train}$ of the training samples are modal-incomplete and  $ 100\% - r_{train}$ are modal-complete. Thus, MMT can:   

1. \textit{Learn from modal-incomplete only.} During training, we use MMT to represent the missing modality inputs for modal-incomplete samples. The tokens of modal-complete samples are never replaced with MMT \ie, $p = 0$.

2. \textit{Learn from modal-complete and modal-incomplete.} Use modal-incomplete samples as in (1), and \textit{random-replace} with non-zero $p$. 

\noindent \textbf{Modal-complete training set.}
As \textbf{all} training samples have complete multimodal inputs, MMT is trained using \textit{random-replace} with $p > 0$.

\noindent\textbf{Inference.} Regardless of the strategy, we replace the missing input tokens with the learned MMT at test time.


\section{Experiments}\label{sec:experiments}

We present a detailed analysis of our MMT under both modal-complete and modal-incomplete training sets. For both scenarios, we explore the usage of \textit{random-replace}. Namely, in Sec.~\ref{subsec:random-replace} we ablate the effect of $p$. In Sec.~\ref{subsec:incomplete}, we study how the severity of missing modality in the training data affects the performance. Sec.~\ref{subsec:both_missing} we extend the setup to multiple missing modalities. Additionally, we study the effect of fusion layers $L_f$ in Sec.~\ref{subsec:fusion_layer}. Then, we compare our method to the baselines we proposed and~\cite{lee2023multimodal} in Sec.~\ref{subsec:sota}.
\subsection{Datasets}
We use videos from \textbf{Ego4D~\cite{grauman2022ego4d}} for pre-training the MBT backbone. Due to privacy and regulations, only 2.5K of 3.7K video hours have original audio in this dataset.  We trim 450K 10-second modal-complete audiovisual clips spanning all 2.5K modal-complete hours. For the downstream tasks, we use the following egocentric action recognition benchmarks: 

\noindent\textbf{Epic-Kitchens-100}~\cite{damen2022rescaling} has 90K trimmed clips of variable length, spanning 100 video hours.  Each clip is labeled with a noun + verb pair, which describes the camera-wearer action. In total, there are 300 noun and 97 verb classes in the dataset. We train the model with 2 heads to jointly predict verb and noun classes. All videos in the dataset have complete visual and audio streams ($r_{train} = r_{test} = 0\%$).  

\noindent\textbf{Epic-Sounds}~\cite{huh2023epic} spans the same 100 video hours as Epic-Kitchens but is annotated with sound labels. This dataset does not follow the noun and verb annotations from Epic-Kitchens; instead, it has 44 unique class labels. The dataset is composed of 79K annotated clips.

\noindent\textbf{Ego4D-AR:} We use the annotated clips from the Short-Term Action Anticipation task in Ego4D benchmark~\cite{grauman2022ego4d} to create an action recognition dataset that we dub as Ego4D-AR\footnote{Ego4D does not have an action recognition benchmark}. Specifically, we use the provided \textit{time-to-contact} timestamps to trim the clips and the anticipated actions as labels. Ego4D-AR contains 142K clips annotated as noun and verb pairs. Overall, there are 128 noun and 81 verb classes. We find that the verb classes are highly imbalanced in this dataset. Therefore, we balance the class weights in the cross-entropy loss during training. We provide more details on the dataset in Supplementary. Similarly to Epic-Kitchens, we use 2 heads to predict the nouns and verbs. As we didn't filter the Ego4D videos for this dataset, it has naturally missing modality. Only $71\%$ of training clips and $73\%$ of test clips have audio ($r_{train} = 29\%$, $r_{test} = 27\%$).

\noindent For clarity and due to space constraints, we report the verb accuracy for Epic-Kitchens, the class accuracy for Epic-Sounds, and the noun accuracy for Ego4D-AR in this section. We report the rest of the metrics for Supplementary.

\subsection{Implementations details}
\noindent\textbf{Pre-training.}
We use the audiovisual MAEs~\cite{he2022masked, feichtenhofer2022masked, georgescu2023audiovisual} protocols and train our own implementation of Audiovisual Bottleneck MAE. We use the trimmed Ego4D clips and train for 200 epochs. We mask 70\% of audio and 90\% of video tokens. We use the same pre-trained model for all experiments. More details of the decoder used in the pre-training can be found in the supplementary material. 

\noindent\textbf{Architecture.}
Following~\cite{huang2022mavil}, we use ViT-Base~\cite{dosovitskiy2020image}  with 12 transformer layers, 12 attention heads, and embedding dimension 768 as the encoder for each modality. For the fusion design, we follow MBT~\cite{nagrani2021attention} and fix the number of bottlenecks to $B = 4$ and the fusion layer to $L_f = 8$ (except for Sec.~\ref{subsec:fusion_layer}). 

\noindent\textbf{Inputs.}
Following~\cite{gong2021ast, huang2022mavil}, we convert an audio waveform of $t$ seconds to log Mel-filterbank with 128 Mel-frequency bins, with a Hanning window of 25ms, shifting every 10ms. The output is a spectrogram of $128 \times 100t$. We use 8-second audio and the patch size of $16\times16$, resulting in $(128\times100\times 8) /256 = 400$ audio tokens. For video, we sample 16 RGB frames at $8$ fps of $224\times224$. Similarly to~\cite{huang2022mavil}, we tokenize the frames with 3D convolutions, using the spacetime patch size of $16\times16\times2$. Each video input produces ($16\times224\times224)/(256\times2) = 1568$ tokens.

\noindent\textbf{Finetuning.}
We train for 50 epochs in Epic-Kitchens experiments, 20 in Epic-Sounds, and 15 in Ego4D-AR. We use SpecAugment~\cite{park2019specaugment} 
for audio augmentation and Augmix~\cite{hendrycks2019augmix} 
for video augmentation. We use AdamW~\cite{loshchilov2016sgdr} optimizer with half-cycle cosine learning rate decay.

\begin{table}[h]
\centering
\small
\caption{\textbf{The performance of the audio, video, and bottleneck audiovisual models on each dataset. } We train and evaluate all models with $r_{train}$ = 0 and $r_{test} = 0\%$. In Ego4D-AR, this is done by filtering out the modal-incomplete samples. In all datasets, multimodal performance beats unimodal performance. }
\begin{tabular}{l|ccc}
\hline
\textbf{Dataset} &  \textbf{Audio} & \textbf{Video} & \textbf{Audiovisual}    \\ \hline 
Epic-Kitchens      & 40.0\%        & 63.2\%        & 64.0\%                      \\
Epic-Sounds        & 46.5\%        & 41.4\%        & 55.2\%                                    \\
Ego4D-AR         & 26.3\%        & 34.6\%        & 36.4\%                      \\ \hline
\end{tabular} 

\label{tab:datasets}
\end{table}
\subsection{Unimodal and baseline multimodal models}
In Table~\ref{tab:datasets}, we show the unimodal and multimodal performance for each downstream dataset. As the original MBT, the multimodal models are trained with fully modal-complete samples. For our \textit{baseline} results across all datasets, we employ these multimodal models. In scenarios where $r_{test}$ is small, indicating that the testing data is nearly modal-complete, it is desired that the adapted model exhibits performance more closely aligned with the multimodal baseline model. Consequently, \textit{this adaptation should uphold the multimodal reasoning capabilities of the model.}

As expected due to the annotation strategy, video serves as the dominant modality in Epic-Kitchens and Ego4d-AR, and audio takes precedence in Epic-Sounds. Consequently, as detailed in Sec.~\ref{subsec:task}, we train our models to be robust to missing video in Epic-Kitchens and Ego4D-AR and to missing audio in Epic-Sounds (except for Sec.~\ref{subsec:both_missing} where we assume that any modality could be absent). Additionally, we report the unimodal performance of the non-missing modality (referred to as \textit{unimodal}, \ie, video in Epic-Sounds) as \textit{we expect the adapted models to converge towards this performance at higher values of  $r_{test}$}. 

Since Epic-Sounds has fewer training samples and exhibits a more balanced unimodal performance across each modality, we leverage it more extensively in Sec.~\ref{subsec:incomplete} and Sec.~\ref{subsec:both_missing}.

\subsection{Results with MMT.}\label{subsec:random-replace}


\begin{figure}[t]
\begin{center}
\includegraphics[width=\linewidth]{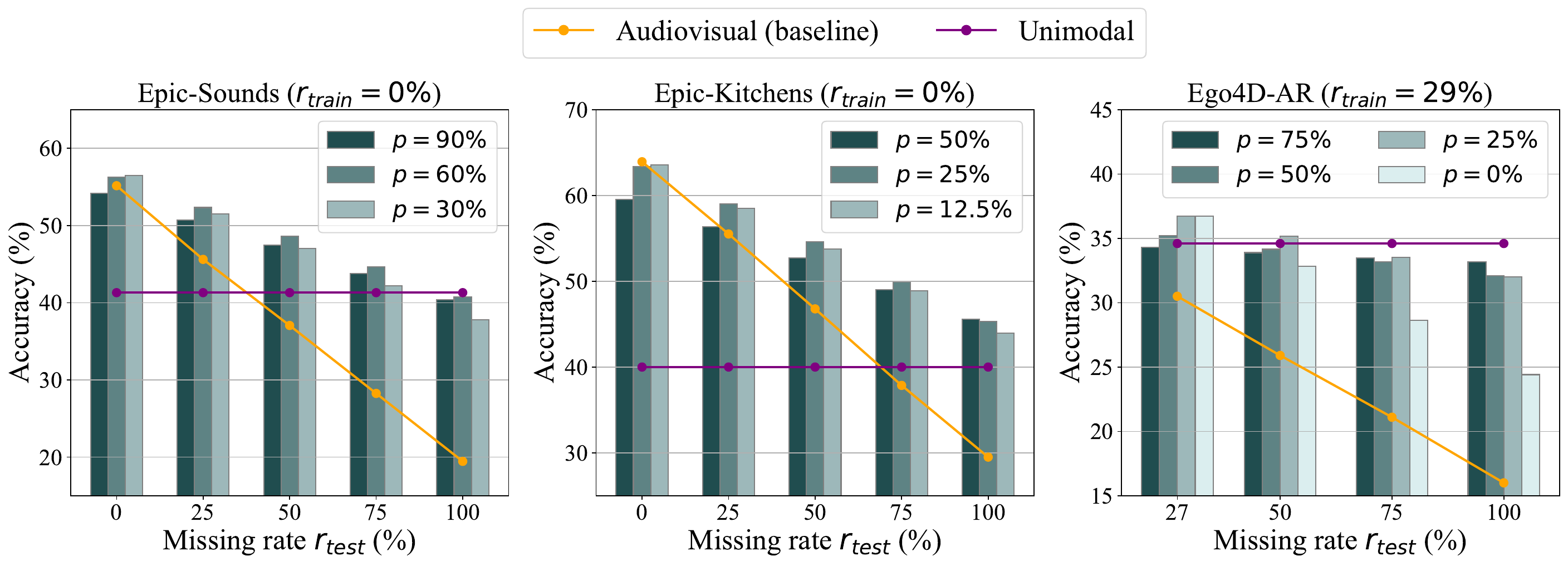}
\end{center}
\small
\vspace{-5pt}
\caption{\textbf{Modality drop probability $p$ \vs accuracy for modal-complete  Epic-Sounds, Epic-Kitchens and Ego4D-AR.} In all datasets, our method dramatically improves the performance of the baseline (orange). 
}
\label{fig:sound-p-ablation}
\vspace{-10pt}
\end{figure}

We apply MMT as mentioned in Sec.~\ref{subsec:ours}. Fig.~\ref{fig:sound-p-ablation} shows changes in performance with different $p$ in \textit{random-replace} and compares it with the unimodal (purple) and baseline multimodal (orange) performance in each dataset. We use  $p \in \{12.5\%, 25\%, 50\%\}$ for Epic-Kitchens. Ego4D-AR has naturally missing modality with $r_{train} = 29\%$, and we ablate with $p \in \{0\%, 25\%, 50\%, 75\%\}$. We observed that higher values of $p$ yield better performance in Epic-Sounds; hence, we use $p \in \{30\%, 60\%, 90\%\}$ in this dataset.

By looking at the model's performance across all $p$ values, we notice the importance of picking $p$ large enough for the model to adapt well but not too large to avoid high information loss. On the one hand, if there are insufficient training samples for MMT (smaller $p$), the model does not perform optimally at higher $r_{test}$. For example, in Epic-Sounds, the model trained with  $p = 30\%$ reaches 42.7\% at $r_{test} = 25\%$ while increasing $p$ to 60\% gives 44.6\%. Similarly, if we only use naturally modal-incomplete samples to train MMT in Ego4D-AR ($p = 0\%$), the model reaches 28.6\% at $r_{test} = 25\%$, while if we increase to $p = 25\%$, the performance increases to 33.5\%. On the other hand, if $p$ is too large, the model performs suboptimally for more modal-complete test data (smaller $r_{test}$). For example, while the model trained with $p = 75\%$ achieves higher accuracy at $r_{test} = 100\%$ in Ego4D-AR, it seems that it learns to ignore the audio completely, as the performance at $r_{test} = 0\%$ drops to the unimodal video performance in this dataset.

 
We find that the models trained with $p = 60\%$, $p = 25\%$, and $p = 25\%$  in Epic-Sounds, Epic-Kitchen, and Ego4D-AR, respectively, yield the best overall performance in each dataset. These models \textit{perform significantly better than the baseline models (orange)} in the missing modality scenarios. For instance, at $r_{test} = 50\%$, the adapted models improve the baseline performance by $11.5\%$ points in Epic-Sounds, $7.8\%$ points in Epic-Kitchens, and $9.3\%$ points in Ego4D-AR.
Furthermore, for Epic-Sound and Epic-Kitchens, the models trained with MMT \textit{reach or surpass the unimodal performance} at extremely severe $r_{test} = 100\%$. Nevertheless, the adapted models maintain the multimodal baseline performance at $r_{test} = 0\%$, showing that \textit{the adaptation strategy does not harm} the model's capabilities.  

Note how in Ego4D-AR, the baseline fails to reach the unimodal accuracy event at the lowest $r_{test} = 27\%$, making our models trained with MMT a significantly better choice.  This happens because $r_{train} = 29\%$ of samples were filtered out to train the multimodal baseline, causing information loss. This demonstrates how MMT enables better leverage of data in modal-incomplete datasets.

In Epic-Kitchens, the model performs better with $p = 25\%$ much smaller than in Epic-Sounds. We believe this is because video (the missing modality) produces almost $4 \times$ more tokens than audio; thus, replacing video tokens causes more information loss.
\begin{figure}[t!]
\begin{center}
\includegraphics[width=\linewidth]{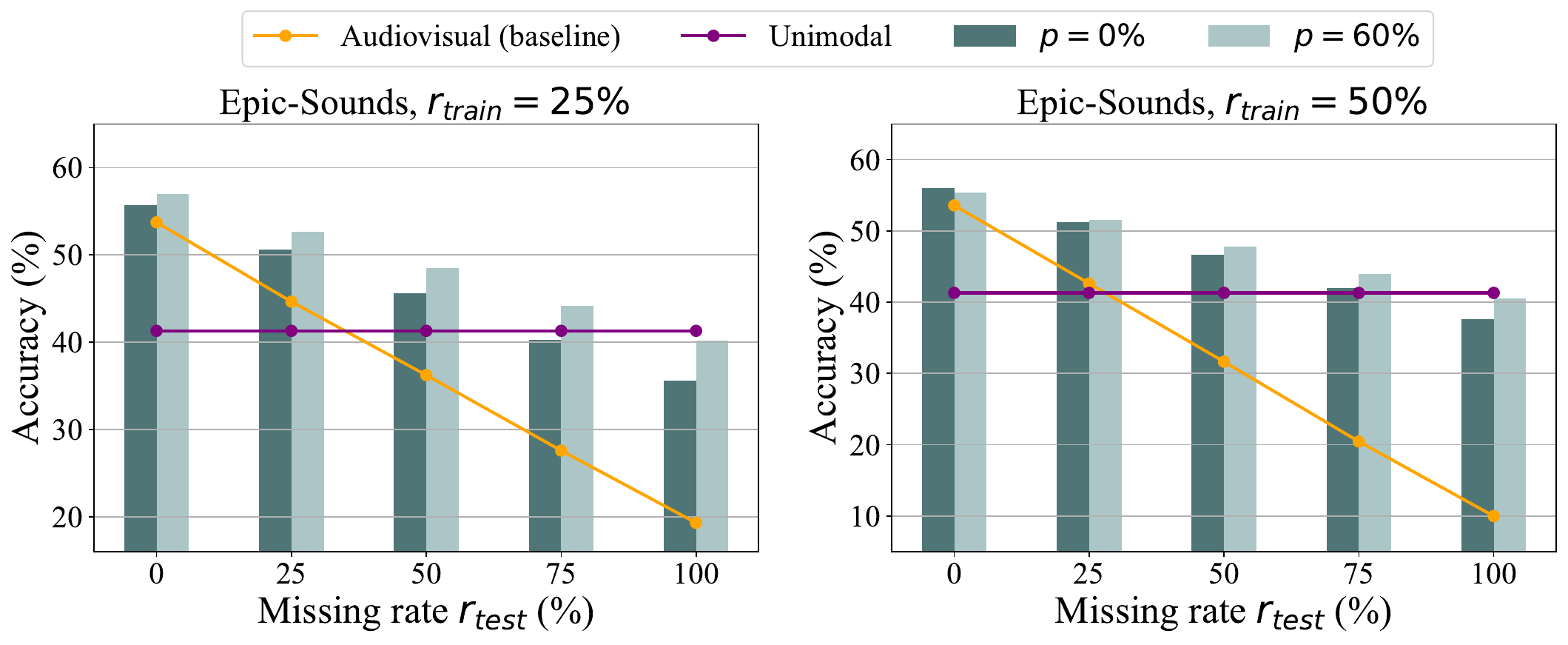}
\end{center}
\small
\caption{\textbf{Results with the modal-incomplete training data.}  As Epic-Sounds does not naturally have missing modality in the training data, we manually remove the audio from (\textbf{left}) $r_{train} = 25\%$  and (\textbf{right}) $r_{train} = 50\%$ of samples in the train set.  
}
\vspace{-10pt}
\label{figure:miss_at_train}
\end{figure}

\subsection{Datasets with modal-incomplete training data.} \label{subsec:incomplete}

In Sec.~\ref{subsec:random-replace}, Ego4D-AR is the only naturally modal-incomplete dataset, and we want to know how our method generalizes in other datasets with different $r_{train}.$ What happens if a dataset has even higher $r_{train}$ than Ego4D-AR? Filtering out noisy or corrupted data causes significant information loss if $r_{train}$ is high. By using MMT, we can mitigate that and learn from all instances. 

We show the results with our modal-incomplete version of Epic-Sounds with $r_{train} = 25\%$ and $r_{train} = 50\%$ in Fig.~\ref{figure:miss_at_train}. We discuss the details of creating this version of the dataset in Supplementary. We can see how, indeed, the baseline model performs suboptimally, similarly as in Ego4D-AR in Sec.~\ref{subsec:random-replace}. Using \textit{random-replace} with $p = 60\%$ significantly improves the baseline accuracy at $r_{test} = 100\%$ by 20 points when $r_{train} = 25\%$ and 30 points when $r_{train} = 50\%$. Furthermore, using modal-complete samples to train MMT ($p = 60\%$) causes a significant performance boost compared to the model trained with modal-incomplete samples only ($p = 0\%$).

\begin{figure}[t!]
\begin{center}
\includegraphics[width=\linewidth]{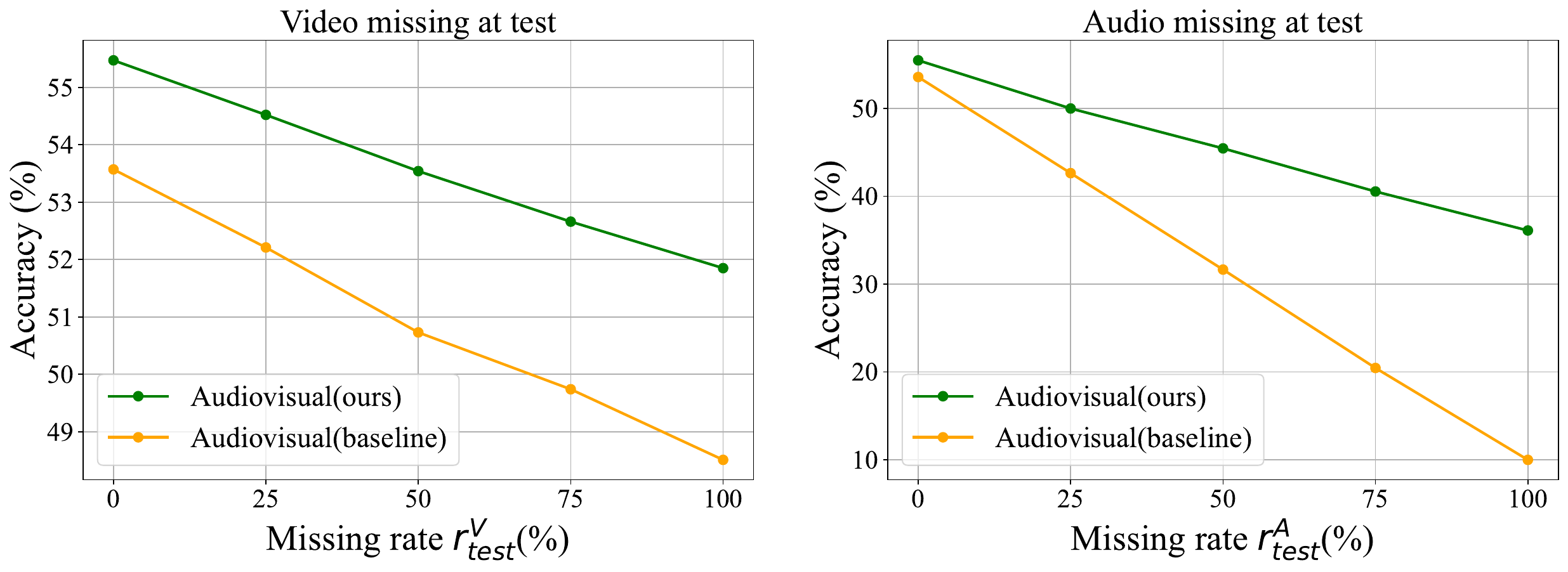}
\end{center}
\small
\caption{Results on Epic-Sounds with $r_{train}^A = 25\%, r_{train}^V = 25\%$. We train our model with two MMTs: one for missing video and one for audio. We run the inference twice: (\textbf{left}) with missing video and (\textbf{right}) missing audio.}
\vspace{-10pt}
\label{figure:bothmodalities}
\end{figure}
\begin{figure*}[t]
\begin{center}
\includegraphics[width=\linewidth]{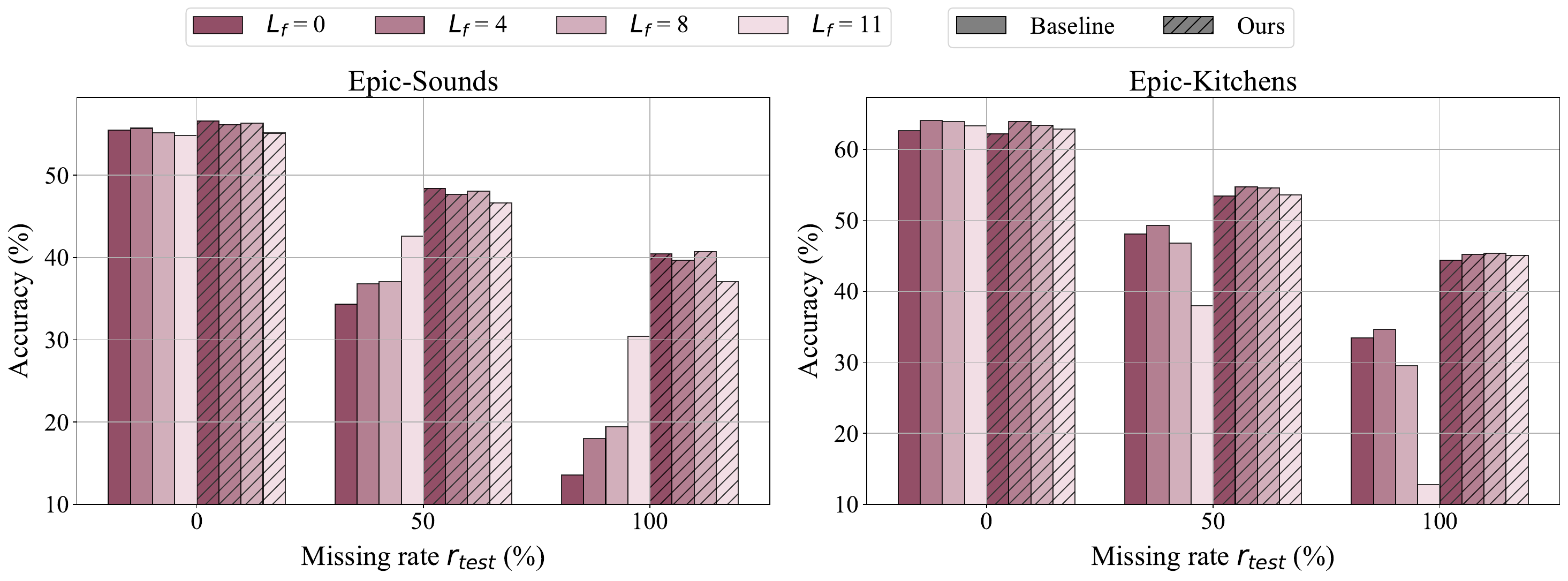}
\end{center}
\small
\caption{\textbf{Fusion Layer $L_f$ \vs accuracy in the models trained with no adaptation strategy (Baseline) and trained with MMT (Ours).} For training with MMT, we use \textit{random-drop} strategy with $p=60\%$ for Epic-Sounds and $p = 25\%$ for Epic-Kitchens. This strategy makes MBT more robust to missing modalities across all $L_f$ and significantly reduces the negative effect of missing modalities.}
\label{figure:fusion}
\end{figure*}



\subsection{Both modalities missing} \label{subsec:both_missing}
In our setup, we train one MMT per dataset because we assume that only one modality could be missing. However, in more realistic scenarios, either modality could be missing for each sample in the dataset. Training a conventional multimodal baseline involves filtering out all modal-incomplete samples, causing the training data to shrink to a small intersection subset where all modalities are available, which could negatively affect the model's performance. 

Luckily, it is quite straightforward to extend our approach to two MMTs, one for each modality.
To simulate this scenario, we use Epic-Sounds and remove audio from $r_{train}^A = 25\%$ \footnote{We use superscripts $^{\{A,V\}}$ to refer to the missing modality} of the training samples and video from another $r_{train}^V= 25\% $ of the training samples. For simplicity, we set $p = 0$ for both MMTs when training our adapted model.
To train a multimodal baseline, we filter out all modal-incomplete samples \ie, half of the training samples.  Fig.~\ref{figure:bothmodalities}, we show how our model's performance compares to the baseline. We evaluate it for missing video (left) and audio (right). As can be seen, training with MMT dramatically improves the model performance. 

\subsection{The effect of the fusion layer.} \label{subsec:fusion_layer}
As we mentioned in Sec.~\ref{subsec:fusion}, the fusion layer does affect the performance of the bottleneck model, especially when test inputs are modal-incomplete. In Fig.~\ref{figure:fusion}, we examine whether training with MMT enhances robustness to a missing modality at various fusion layers and whether models trained with MMT (ours) show the same sensitivity to the fusion layer as those trained without it (baseline). As we observe, \textbf{the introduction of MMT makes the model more robust to the missing modality across all fusion layers} in both Epic-Sounds and Epic-Kitchens. With extremely severe $r_{test} = 100\%$, the adapted models perform with $\sim 45\%$ accuracy in Epic-Kitchens, while the unimodal audio model achieves $40\%$ in this dataset.  Furthermore, these adapted models do not exhibit similar sensitivity to the fusion layer as the baseline. For example, in Epic-Sounds, the baseline models trained with $L_f = 11$ exhibit superior performance, which is not the case for the adapted models (at $r_{test} = 100\%$, $37\%$ accuracy with $L_f = 11$ but $\sim 40\%$ for other $L_f$). Overall, all adapted models perform consistently well, each providing decent performance in a modal-incomplete inference. Our approach effectively addresses a long-standing issue~\cite{ma2022multimodal} of selecting the appropriate fusion layer when faced with missing modalities.

 \subsection{MMT \vs other missing modality representations}\label{subsec:sota}
 In Tables~\ref{tab:combined}, we report the results using our \textit{random-replace} and the baselines from Sec.~\ref{subsec:baseline} for Epic-Sounds, Epic-Kitchens, and Ego4D-AR. We also report the accuracy of the unimodal model in each dataset. Interestingly, passing zeroes as missing modality works better for Epic-Sounds and skipping missing modality tokens works better for Epic-Kitchens and Ego4D-AR.
 
 To compare to the baseline tailored to the missing modality problem, we implement the multimodal prompts method~\cite{lee2023multimodal}. It was originally based on ViLT~\cite{kim2021vilt} for image-text classification, so we implemented our version for action recognition based on MBT. Note that this method relies on a strong pre-trained backbone to efficiently finetune it by optimizing a few network parameters. As we deal with audiovisual learning in egocentric videos, we rely on large-scale egocentric pre-training using MAEs. As seen in Tab.~\ref{tab:combined}, our method outperforms the one proposed by~\cite{lee2023multimodal}.  
\begin{table}[h!]
\centering
\small
\caption{\textbf{Comparison of our method with baselines in Epic-Sounds, Epic-Kitchnes, and Ego4D-AR.} We demonstrate the accuracy across different missing modality ratios $r_{test}$. We show in \textbf{bold} the best result and \underline{underline} the runner-up. We mentioned the way of representing the missing modality in brackets: \textit{zeros} for the baseline of passing zeros,  \textit{skip} for the baseline skipping the tokens of the missing input, as mentioned in Sec.~\ref{subsec:baseline}, or MMT. }
\begin{tabular}{l|c|c|c|c|c|c}
\toprule

\multirow{2}{*}{Dataset} & \multirow{2}{*}{$r_{test}$} & \textcolor{gray}{Unimodal} & Baseline & Baseline & Prompts~\cite{lee2023multimodal} & Ours \\
 & & & (zeros) & (skip) & & (MMT) \\
\hline
\multirow{5}{*}{Epic-Sounds} & 0\% & \textcolor{gray}{41.4} & \underline{55.2} & \underline{55.2} & 36.7 & \textbf{56.3} \\
 & 25\% & \textcolor{gray}{41.4} & 45.6 & \underline{47.0} & 33.3 & \textbf{52.3} \\
 & 50\% & \textcolor{gray}{41.4} & 37.1 & \underline{39.9} & 30.0 & \textbf{48.6} \\
 & 75\% & \textcolor{gray}{41.4} & 28.3 & \underline{32.5} & 26.1 & \textbf{44.6} \\
 & 100\% & \textcolor{gray}{41.4} & 19.5 & \underline{25.0} & 22.6 & \textbf{40.7} \\
\hline
\multirow{5}{*}{Epic-Kitchens} & 0\% & 
         \textcolor{gray}{40.0} & \textbf{63.9} &	\textbf{63.9} &	32.5 &	63.4 \\
& 25\% & \textcolor{gray}{40.0} & \underline{55.5} & 53.2  & 31.1 &	\textbf{59.0} \\
& 50\% & \textcolor{gray}{40.0} & \underline{46.8} & 42.1 &  30.2 & \textbf{54.6} \\
& 75\% & \textcolor{gray}{40.0} & \underline{37.9} & 30.8 & 29.2 & \textbf{50.0} \\
& 100\% & \textcolor{gray}{40.0} & \underline{29.5} & 20.0 &  28.2 & \textbf{45.3} \\
\hline
\multirow{4}{*}{Ego4D-AR}  & 27\% & \textcolor{gray}{34.6} &	30.5 & \underline{32.5}	& 18.4 &	\textbf{36.7 }\\
& 50\%  & \textcolor{gray}{34.6} & 25.9 & \underline{30.8} &16.4&	\textbf{35.2} \\
& 75\%  & \textcolor{gray}{34.6} & 21.1 & \underline{29.1} & 13.9	& \textbf{33.5} \\
& 100\% & \textcolor{gray}{34.6} & 16.0 & \underline{27.2} &11.5	& \textbf{32.0} \\
\bottomrule
\end{tabular}
\label{tab:combined}

\end{table}

\noindent We find that across all datasets, our MMT trained with \textit{random-replace} either reach (in Epic-Sounds and Ego4D-AR) or exceed (in Epic-Kitchens) the unimodal performance in extreme $r_{test} = 100\%$. Interestingly, in Epic-Sounds, \textit{random-replace} also regularizes the training and increases the $r_{test}=0\%$ performance by 1.1 points.

\section{Conclusion}
We explore the missing modality problem in multimodal egocentric datasets. We suggest a simple yet effective method by learning the optimal token representation of the missing modality (MMT). Placing learnable tokens to represent missing inputs provides an easy and intuitive way to train and test with modal-incomplete inputs. We propose strategy \textit{random-replace} to learn MMT when training action recognition models and show how their performance brings us closer to robust and effective multimodal systems.


%
%
\bibliographystyle{splncs04}
\bibliography{main}
\end{document}


\maketitle
\begin{figure*}[h]
\begin{center}
\includegraphics[width=\linewidth]{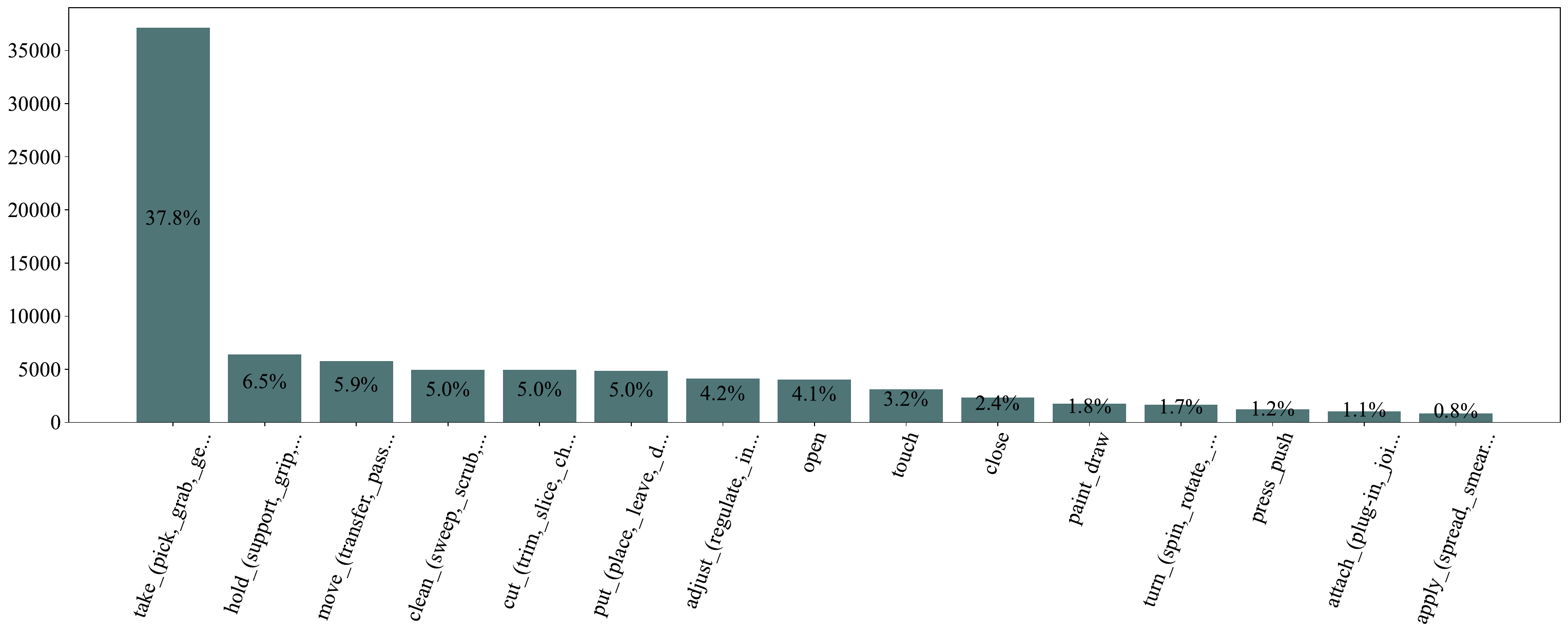}
\end{center}
\small
\caption{\textbf{Distribution of the verb classes in Ego4D-AR.} }
\label{figure:supp-ego4d-class}
\end{figure*}

\begin{figure*}[t!]
\begin{center}
\includegraphics[width=\linewidth]{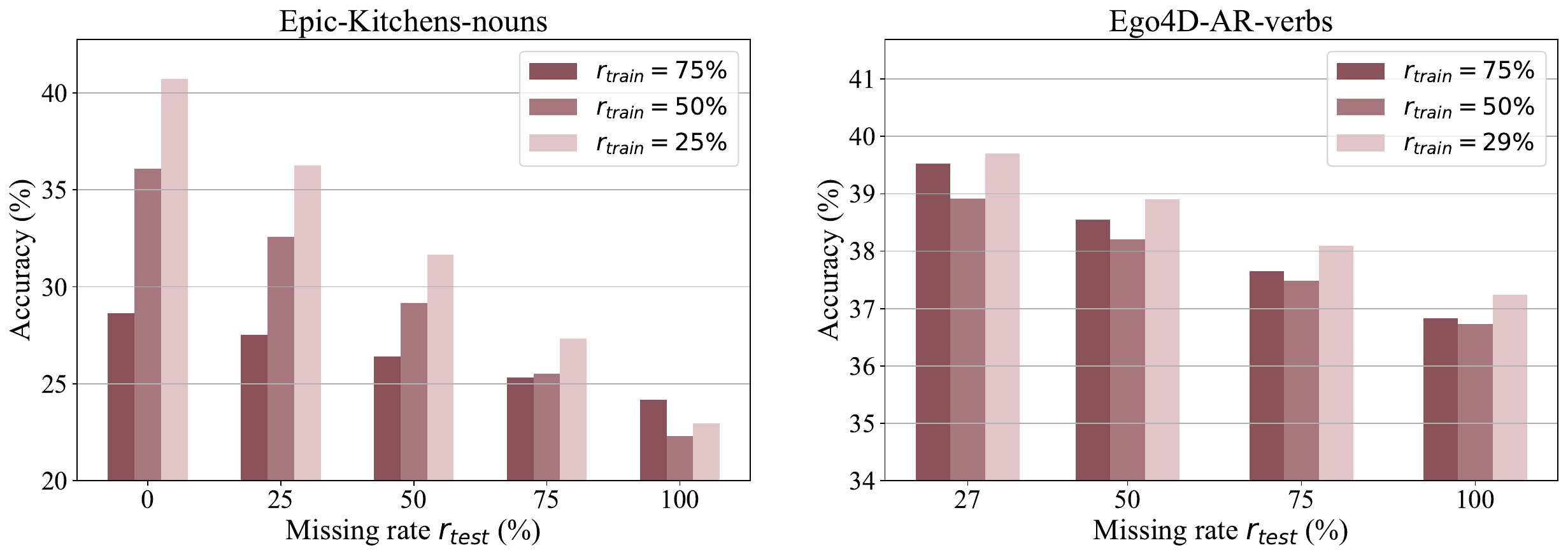}
\end{center}
\caption{\textbf{Missing modality rate in training data $r_{train}$ \vs accuracy when training with \textit{fixed-replace} strategy.} We show that our MMT effectively learns with missing modalities under several $r_{train}$ regimes. For Epic-Kitchens nouns, lower masking yields better accuracy overall. This result is also consistent with to verb classes (main manuscript). Also, higher $r_{train}$ provides better robustness to missing modality at $r_{test} = 100\%$, which is what we observed from the experiments on all datasets.  For Ego4d-AR verbs, there is no obvious correlation between the number of modal-incomplete training samples and the accuracy. As mentioned in Section~\ref{supp-section:dataset}, this is due to the highly imbalance distribution of verb classes. 
}
\label{figure:supp-fixed}
\end{figure*}

\input{figures_supplementary/random-epic}
\begin{figure}[h]
\begin{center}
\includegraphics[width=\linewidth]{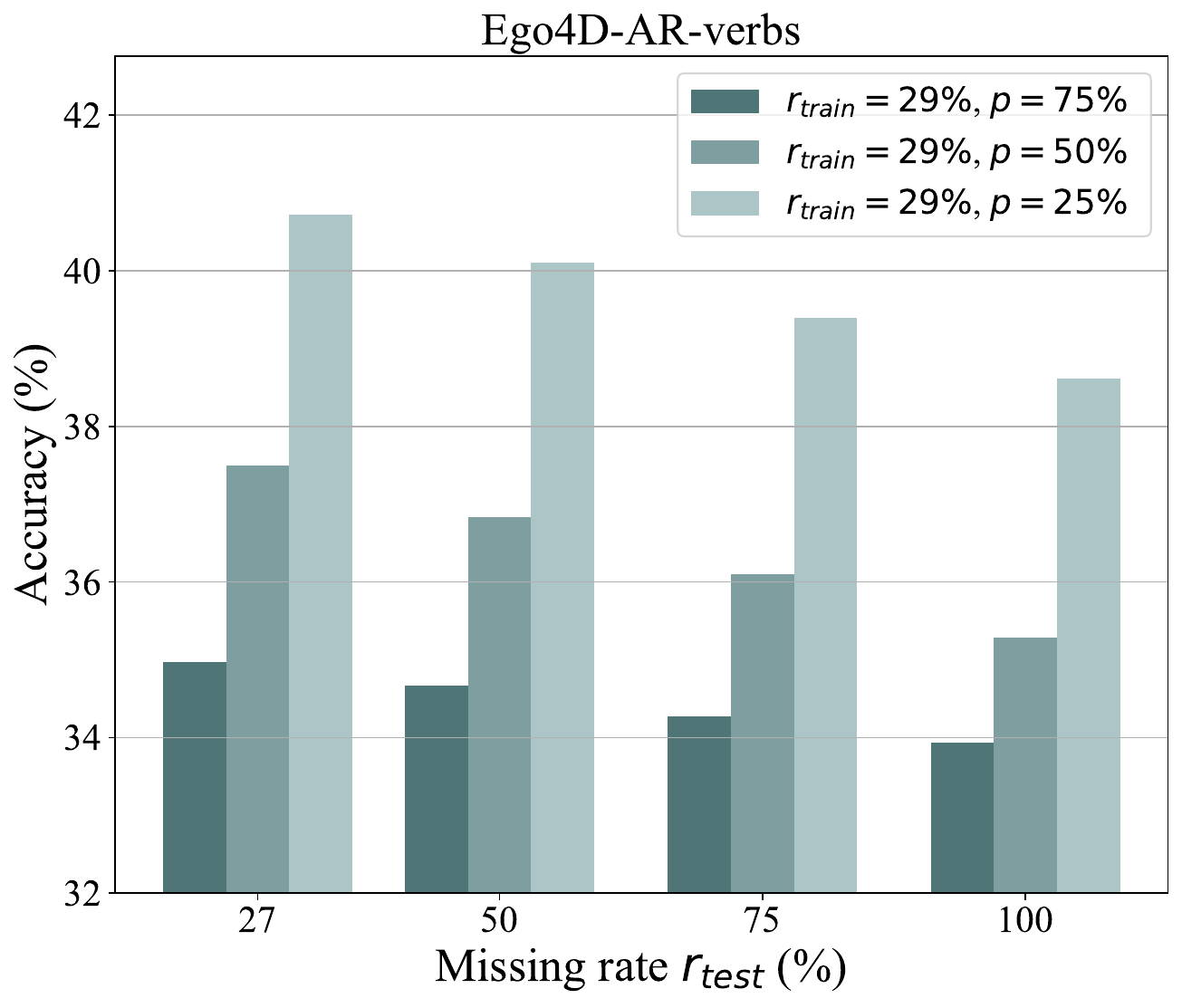}
\end{center}
\small
\caption{\textbf{ Modality drop probability $p$ \vs accuracy for modal-incomplete Ego4D-AR}. We can observe the overall performance drop with more extreme modality replacement with MMT $r_{train} = 29\%, p = 50\%$ and  $r_{train} = 29\%, p = 75\%$, which is consistent with the noun performance. However, we find that the models overall are not trained well due to overfitting to the imbalanced class distribution, and the effect of MMT is hard to observe.} 
\label{fig:supp-random-ego4d}
\end{figure}

\begin{wrapfigure}{r}{0.55\textwidth}
\resizebox{0.55\textwidth}{!}{
\begin{minipage}{\linewidth}
\begin{center}
\includegraphics[width=\linewidth]{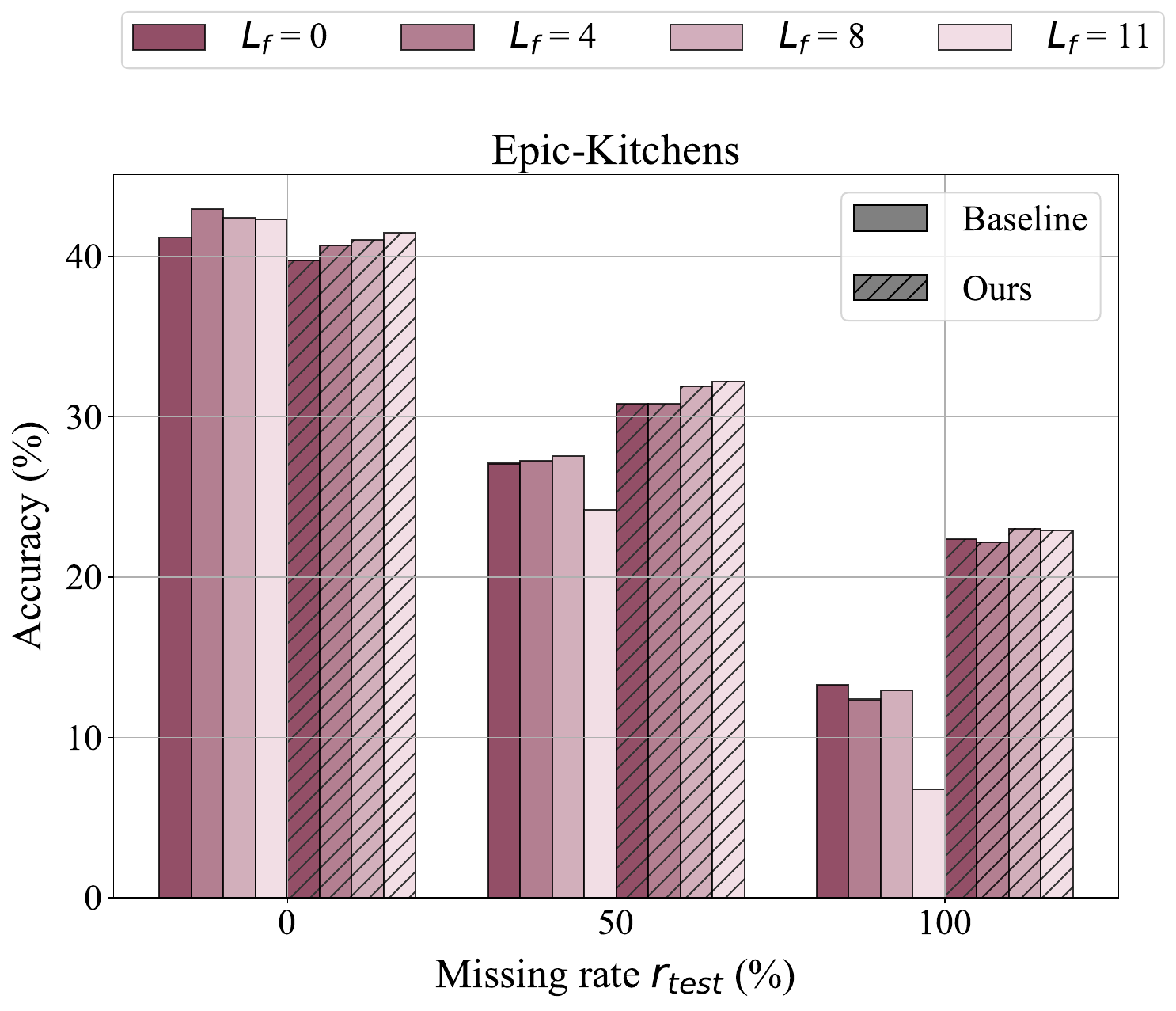}
\end{center}
\small
\caption{\textbf{Fusion Layer $L_f$ \vs accuracy in the models trained with no adaptation strategy (Baseline) and trained with MMT (Ours).} For training with MMT, we use \textit{random-drop} strategy with $p = 25\%$ for Epic-Kitchens. Our strategy makes the model more robust to missing modalities across all $L_f$.}
\label{figure:supp-fusion}
\end{minipage}
}
\vspace{-2.0cm}
\end{wrapfigure}

\section{Ego4D-AR}
As mentioned in the main manuscript, we curate action recognition clips from the Ego4D Short-term Action Anticipation benchmark~\cite{grauman2022ego4d}. Given a video clip, Short-term Action Anticipation predicts the possible future object interaction and a "time to contact" estimate. The future object interaction is predicted as a noun and verb pair, defining the object and the type of interaction. The "time to contact" is the number of seconds after which the interaction is expected to happen. Given these annotations, we trim the clips centered around the ground-truth "time to contact" and assign them the ground-truth noun and verb annotations. Following that, we obtained 98276 train and 47395 test instances for the action recognition dataset Ego4D-AR.

We observed that the dataset's class distribution is highly imbalanced. To demonstrate this, we plotted the distribution of the top-15 most common verb classes in the training set (see Figure~\ref{figure:supp-ego4d-class}). Notably, over 35,000 instances, or 38\%, belong to the verb class `take'. Training with this imbalance is challenging as the model tends to become biased towards this verb class. To address this, we adapted a common strategy: adjusting per-class weights in the cross-entropy loss for verb prediction. Let \( |S_i| \) denote the number of samples in the dataset labeled as class \( i \), where \( S_i = \{ x \in \text{Dataset} \mid \text{label}(x) = i \} \). The weight for class \( i \) is then \( w_i = 1 - \frac{|S_i|}{|S|} \), meaning the loss is reduced for more common verb classes and increased for rarer ones. We did not apply this adjustment to noun prediction, as noun label distribution is more balanced. The adjustment of per-class weights in the loss function only slightly mitigated bias in the verb distribution, leading to variable test performance. The notable fluctuations in verb accuracy stem largely from the class imbalance in the dataset. This imbalance presents a significant learning challenge, resulting in inconsistent verb prediction outcomes. Understanding this variability is essential for interpreting the results in the Ego4D-AR experiments.

\label{supp-section:dataset}

\begin{table}[h]
\centering
\small
\caption{Comparison of our method with baselines on various datasets. We demonstrate the accuracy of our method \vs the adaptation of the method proposed by~\cite{lee2023multimodal} and unimodal performance across different missing modality ratios $r_{test}$. We show in \textbf{bold} the best result and \underline{underline} the runner-up.}
\subfloat[\textbf{Epic-Kitchens-nouns} ]{
\begin{tabular}{l|c|c|c|c|c}
 $r_{test}$ & 0\% & 25\% & 50\% & 75\% & 100\% \\
 \hline
\textcolor{gray}{Unimodal} & \textcolor{gray}{19.5}  & \textcolor{gray}{19.5}  & \textcolor{gray}{19.5}  & \textcolor{gray}{19.5}  & \textcolor{gray}{\underline{19.5}}  \\
\hline
Baseline(zeros) & \textbf{42.4} & \underline{35.2} & \underline{27.5} & \underline{20.2} & \underline{12.4} \\ 
Baseline(skip) & \textbf{42.4} & 32.9 & 22.9 & 13.1 & 3.5  \\
Prompts~\cite{lee2023multimodal} & 10.9 & 10.1 & 9.3 & 8.4 & 7.6 \\ 
\hline
\rowcolor[gray]{0.9} Ours (MMT) & \underline{41.0} & \textbf{36.6} & \textbf{31.9} & \textbf{27.6} & \textbf{23.0}  \\ 

\hline
\end{tabular}
\label{tab:epic_main}
}
\subfloat[\textbf{Ego4D-AR-verbs} ]{
\begin{tabular}{l|c|c|c|c}
 $r_{test}$ &  27\% & 50\% & 75\% & 100\% \\
 \hline
\textcolor{gray}{Unimodal} & \textcolor{gray}{36.5} & \textcolor{gray}{36.5} & \textcolor{gray}{36.5} & 36.5  \\
\hline
Baseline(zeros) & 34.8 & 30.6 & 25.9 & 21.4 \\ 
Baseline(skip) &\underline{39.5} & 37.8 & 36.1 & 34.5 \\
Prompts~\cite{lee2023multimodal} &  38.8 & \underline{38.8} & \underline{38.7} & \textbf{38.7}  \\ 
\hline
\rowcolor[gray]{0.9}Ours (MMT) &  \textbf{40.7} & \textbf{40.1} & \textbf{39.4} & \underline{38.6} \\ 
\hline
\end{tabular}
\label{tab:ego4d_main}
}

\label{tab:supp-combined}
\end{table}

\section{Accuracy on Nouns in Epic-Kitchens and verbs in Ego4D-AR}
\label{supp-section:accuracy}

For clarity and space constraints, we report the noun class accuracy for Ego4D-AR and verb class accuracy for Epic-Kitchens in the main manuscript. In this section, we provide the results for verbs in Ego4D-AR and nouns in Epic-Kitchens.
\subsection{\textit{fixed-replace}  and \textit{random-replace}}
We report the results similarly to the main manuscript. In Figure~\ref{figure:supp-fixed} we show the results using strategy \textit{fixed-replace}  and in Figures~\ref{fig:supp-random-epic} and~\ref{fig:supp-random-ego4d} - using \textit{random-replace} for both datasets. We find that analyzing the results of Ego4D-AR is hard due to the learning issue in the verbs prediction. However, we find that for Epic-Kitchens, the observations are consistent with the ones reported in the main manuscript: using smaller $r_{train}$ and $p$ yields the best performance. Furthermore, we observe that using \textit{random-replace} achieves slightly superior performance for higher $r_{test}$, when compared to \textit{fixed-replace}. We find that this is how \textit{random-drop } benefits from learning from random samples at each iteration; therefore, allowing the model to observe the same sample as model-complete and modal-incomplete (as we also mentioned in the main manuscript). 

\subsection{Fusion layer}
In Figure~\ref{figure:supp-fusion} we report the results with different fusion layers on Epic-Kitchens nouns. As we observe, the introduction of MMTs makes the model more robust to the missing modality across all fusion layers.
\subsection{Other methods and Ours}
In Table~\ref{tab:supp-combined} we report the results with our methods \vs the baselines. We find that in Epic-Kitchens, our strategies perform much better that the proposed baselines, and outperform the unimodal $19.5\%$ by $3.5\%$ points. As the models suffer from overfitting in Ego4D-AR verbs, we find the results hard to compare. For example, while using MM-prompts~\cite{lee2023multimodal}, the model predicts with the same verb accuracy across all $r_{test}$, demonstrating severe overfitting. 

\section{Decoder in the pre-training}
We are using Masked-Autoencoder (MAE)~\cite{he2022masked} for self-supervised pre-training of the MBT backbone. MAE has two parts: encoder and decoder. During training, some input tokens are masked, and only the non-masked tokens are passed to the encoder.  Then, learnable masked tokens with shared parameters are appended to the encoded non-masked tokens, and all are passed to the lightweight decoder for reconstructing the original input of the masked token. The reconstruction loss is applied to the masked tokens. After the pre-training, the decoder is discarded, and only the encoder is used for the fine-tuning. We provide the encoder details in the main paper and share the decoder details here.

Note that the encoder architecture uses bottleneck fusion, meaning that the two modalities can communicate through several learnable tokens. We do not model the multimodal fusion in the decoder part, so each modality has a separate transformer in the decoder. Each decoder is a 4-layer transformer with 16 attention heads and an embedding dimension of 512.  We do not share the parameters of the transformers (same as in the encoder).
\newpage
{
    \small
    \bibliographystyle{ieeenat_fullname}
    \bibliography{main}
}


\maketitle

\section{Ego4D-AR}
\label{supp-section:dataset}

As mentioned in the main manuscript, we curate action recognition clips from the Ego4D Short-term Action Anticipation benchmark~\cite{grauman2022ego4d}. Given a video clip, Short-term Action Anticipation predicts the possible future object interaction and a "time to contact" estimate. The future object interaction is predicted as a noun and verb pair, defining the object and the type of interaction. The "time to contact" is the number of seconds after which the interaction is expected. Given these annotations, we trim the clips centered around the ground-truth "time to contact" and assign them the ground-truth noun and verb annotations. Following that, we obtained 98276 train and 47395 test instances for the action recognition dataset Ego4D-AR.
\begin{figure*}[h]
\begin{center}
\includegraphics[width=\linewidth]{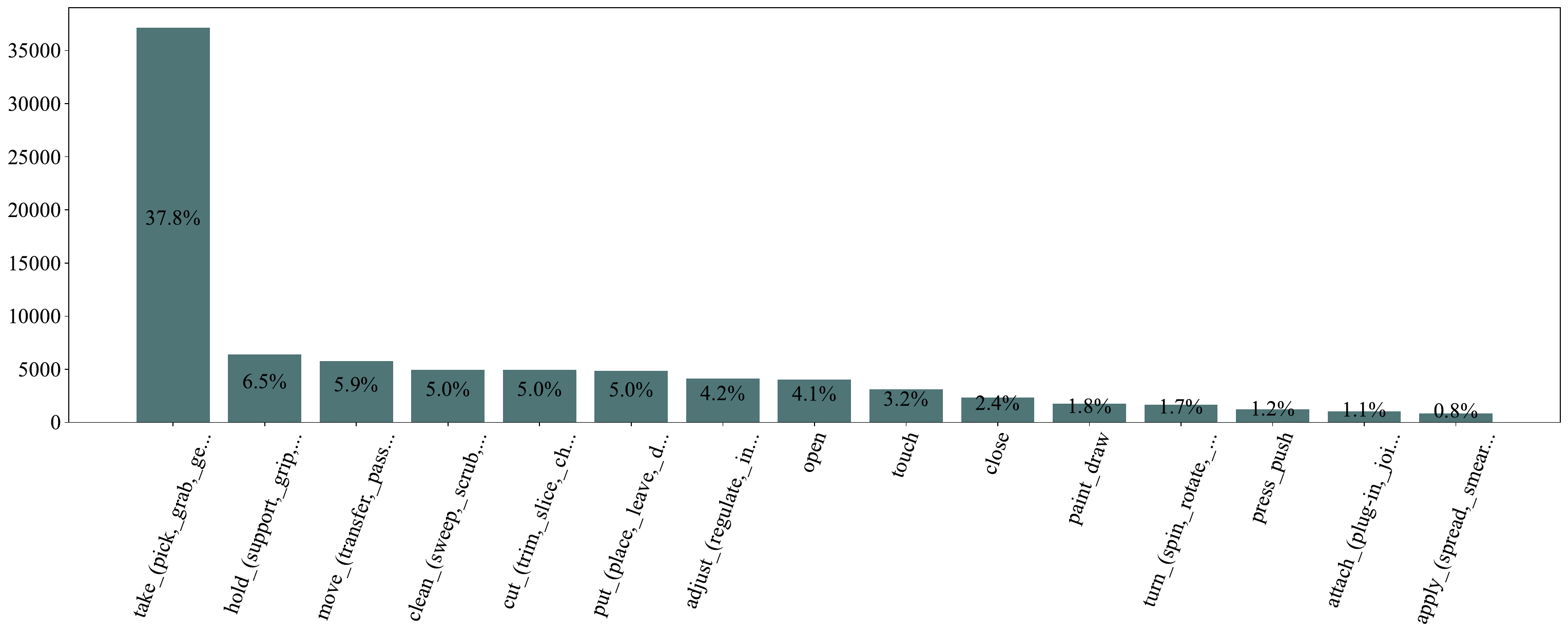}
\end{center}
\small
\caption{\textbf{Distribution of the verb classes in Ego4D-AR.} }
\label{figure:supp-ego4d-class}
\end{figure*}

We observed that the dataset's class distribution is highly imbalanced. To demonstrate this, we plotted the distribution of the top-15 most common verb classes in the training set (see Figure~\ref{figure:supp-ego4d-class}). Notably, over 35,000 instances, or 38\%, belong to the verb class `take'. Training with this imbalance is challenging as the model tends to become biased towards this verb class. To address this, we adapted a common strategy: adjusting per-class weights in the cross-entropy loss for verb prediction. Let \( |S_i| \) denote the number of samples in the dataset labeled as class \( i \), where \( S_i = \{ x \in \text{Dataset} \mid \text{label}(x) = i \} \). The weight for class \( i \) is then \( w_i = 1 - \frac{|S_i|}{|S|} \), meaning the loss is reduced for more common verb classes and increased for rarer ones. We did not apply this adjustment to noun prediction, as noun label distribution is more balanced. Adjusting per-class weights in the loss function only slightly mitigated bias in the verb distribution. The imbalance still presents a significant learning challenge, resulting in high bias in the model predictions. Understanding this is essential for interpreting the results in the Ego4D-AR experiments.

\section{Accuracy on nouns in Epic-Kitchens and verbs in Ego4D-AR}

For clarity and space constraints, we report the noun class accuracy for Ego4D-AR and verb class accuracy for Epic-Kitchens in the main manuscript. In this section, we provide the results for verbs in Ego4D-AR and nouns in Epic-Kitchens.

\subsection{Ablating with different values of $p$ in \textit{random-replace}}
\label{supp-section:accuracy}
\begin{figure}[h]
\begin{center}
\includegraphics[width=\linewidth]{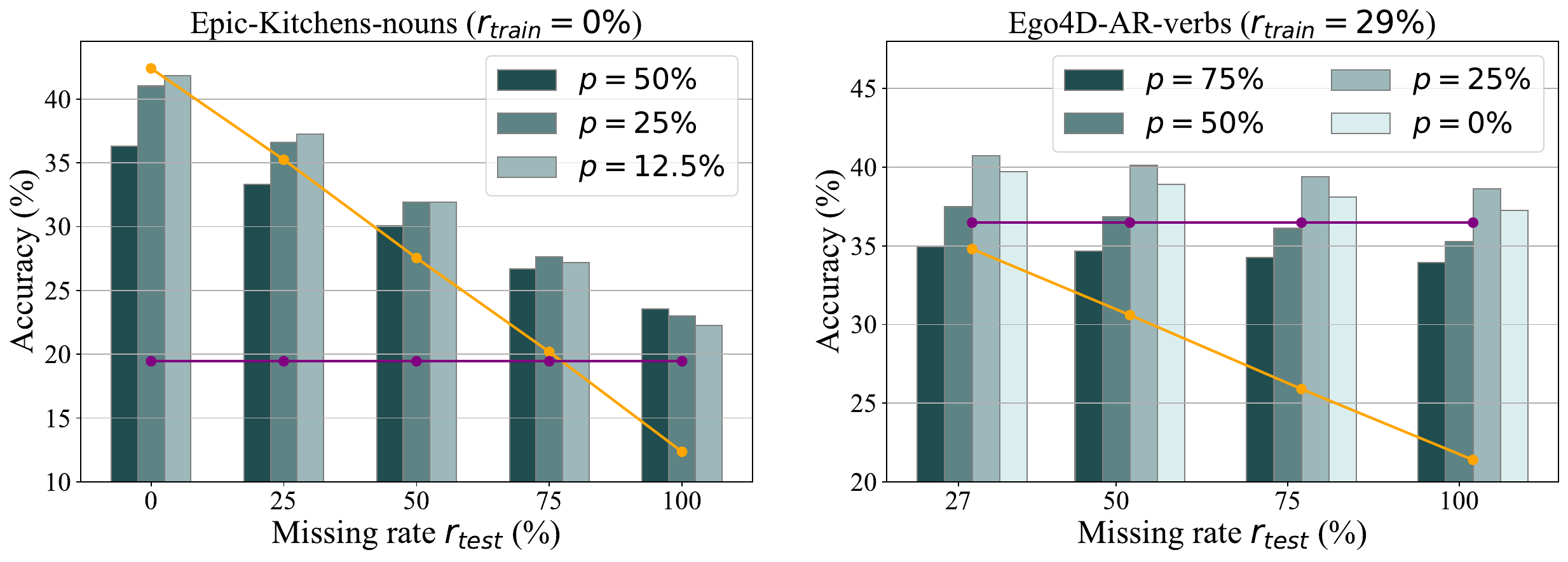}
\end{center}
\small
\caption{\textbf{Modality drop probability $p$ \vs accuracy for modal-complete Epic-Kitchens (nouns) and modal-incomplete Ego4D-AR (verbs).}  
Our MMT effectively learns with missing modalities under several $p$.  The results are consistent with verb classes in Epic-Kitchens and noun classes in Ego4D-AR (main manuscript), \ie smaller $p$ is better for Epic-Kitchens and $p = 25\%$ works the best in Ego4D-AR. For Ego4d-AR verbs, the accuracy is almost the same across all $r_{test}$ for each experiment. As mentioned in Section~\ref{supp-section:dataset}, this is due to the high bias caused by the highly imbalanced distribution of verb classes. 
}
\label{fig:supp-random}
\end{figure}

We report the results similarly to the main manuscript. In Figure~\ref{fig:supp-random}, we show the results using strategy \textit{random-replace} for both datasets. We find that analyzing the results of Ego4D-AR verbs is hard due to the learning issue in the verb prediction. However, we find that for Epic-Kitchens, the observations are consistent with the ones reported in the main manuscript. Selecting the appropriate value of $p$ for each dataset (e.g., $p = 25\%$) is crucial to ensure satisfactory performance across all $r_{test}$ instances. This balance allows the model to effectively adapt while still leveraging the benefits of multimodal information. Furthermore, our models significantly outperform the baseline. 

\begin{wrapfigure}{r}{0.55\textwidth}
\resizebox{0.55\textwidth}{!}{
\begin{minipage}{\linewidth}
\begin{center}
\includegraphics[width=\linewidth]{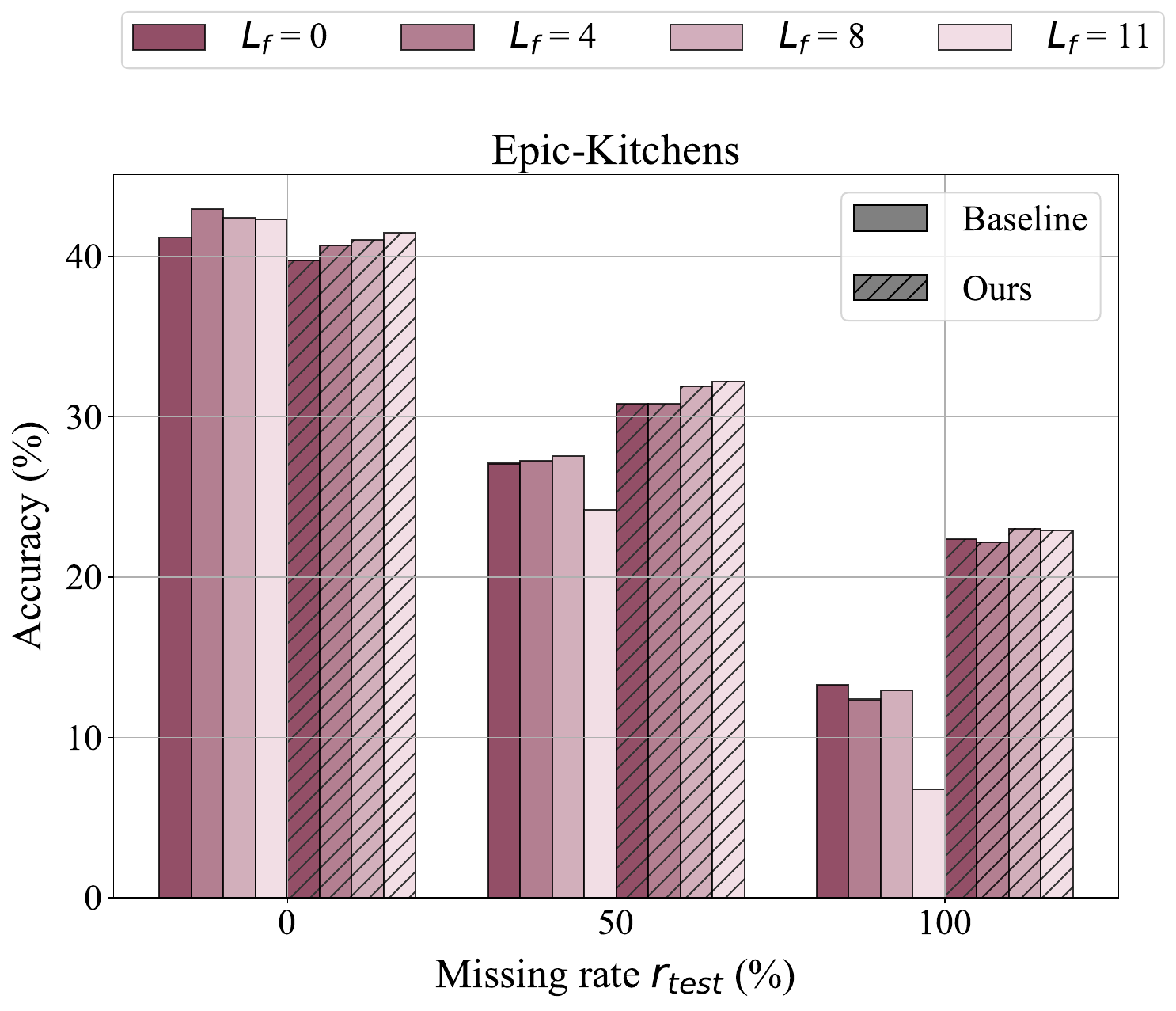}
\end{center}
\small
\caption{\textbf{Fusion Layer $L_f$ \vs accuracy in the models trained with no adaptation strategy (Baseline) and trained with MMT (Ours).} For training with MMT, we use \textit{random-drop} strategy with $p = 25\%$ for Epic-Kitchens. Our strategy makes the model more robust to missing modalities across all $L_f$.}
\label{figure:supp-fusion}
\end{minipage}
}
\vspace{-2.0cm}
\end{wrapfigure}

\subsection{Fusion layer}
In Figure~\ref{figure:supp-fusion}, we report the results with different fusion layers on Epic-Kitchens nouns. As we observe, the introduction of MMTs makes the model more robust to the missing modality across all fusion layers.
\subsection{Other methods and Ours}
In Table~\ref{tab:supp-combined}, we report the results with our methods \vs the baselines. In Epic-Kitchens, our strategies perform much better than the proposed baselines and outperform the unimodal $19.5\%$ by $3.5\%$ points. As the models suffer from high bias in Ego4D-AR verbs, the results are hard to compare. For example, while using prompts~\cite{lee2023multimodal}, the model predicts with the same verb accuracy across all $r_{test}$, demonstrating high bias.

\begin{table}[h]
\centering
\small
\caption{Comparison of our method with baselines on various datasets. We demonstrate the accuracy of our method \vs the adaptation of the method proposed by~\cite{lee2023multimodal} and unimodal performance across different missing modality ratios $r_{test}$. We show in \textbf{bold} the best result and \underline{underline} the runner-up.}
\subfloat[\textbf{Epic-Kitchens-nouns} ]{
\begin{tabular}{l|c|c|c|c|c}
 $r_{test}$ & 0\% & 25\% & 50\% & 75\% & 100\% \\
 \hline
\textcolor{gray}{Unimodal} & \textcolor{gray}{19.5}  & \textcolor{gray}{19.5}  & \textcolor{gray}{19.5}  & \textcolor{gray}{19.5}  & \textcolor{gray}{\underline{19.5}}  \\
\hline
Baseline(zeros) & \textbf{42.4} & \underline{35.2} & \underline{27.5} & \underline{20.2} & \underline{12.4} \\ 
Baseline(skip) & \textbf{42.4} & 32.9 & 22.9 & 13.1 & 3.5  \\
Prompts~\cite{lee2023multimodal} & 10.9 & 10.1 & 9.3 & 8.4 & 7.6 \\ 
\hline
\rowcolor[gray]{0.9} Ours (MMT) & \underline{41.0} & \textbf{36.6} & \textbf{31.9} & \textbf{27.6} & \textbf{23.0}  \\ 

\hline
\end{tabular}
\label{tab:epic_main}
}
\subfloat[\textbf{Ego4D-AR-verbs} ]{
\begin{tabular}{l|c|c|c|c}
 $r_{test}$ &  27\% & 50\% & 75\% & 100\% \\
 \hline
\textcolor{gray}{Unimodal} & \textcolor{gray}{36.5} & \textcolor{gray}{36.5} & \textcolor{gray}{36.5} & 36.5  \\
\hline
Baseline(zeros) & 34.8 & 30.6 & 25.9 & 21.4 \\ 
Baseline(skip) &\underline{39.5} & 37.8 & 36.1 & 34.5 \\
Prompts~\cite{lee2023multimodal} &  38.8 & \underline{38.8} & \underline{38.7} & \textbf{38.7}  \\ 
\hline
\rowcolor[gray]{0.9}Ours (MMT) &  \textbf{40.7} & \textbf{40.1} & \textbf{39.4} & \underline{38.6} \\ 
\hline
\end{tabular}
\label{tab:ego4d_main}
}

\label{tab:supp-combined}
\end{table}

\section{Decoder in the pre-training}
We are using Masked-Autoencoder (MAE)~\cite{he2022masked} for self-supervised pre-training of the MBT backbone. MAE has two parts: encoder and decoder. During training, some input tokens are masked, and only the non-masked tokens are passed to the encoder.  Then, learnable masked tokens with shared parameters are appended to the encoded non-masked tokens, and all are passed to the lightweight decoder for reconstructing the original input of the masked token. The reconstruction loss is applied to the masked tokens. After the pre-training, the decoder is discarded, and only the encoder is used for fine-tuning. We provide the encoder details in the main paper and share the decoder details here.

Note that the encoder architecture uses bottleneck fusion, meaning the two modalities can communicate through several learnable tokens. We do not model the multimodal fusion in the decoder part, so each modality has a separate transformer in the decoder. Each decoder is a 4-layer transformer with 16 attention heads and an embedding dimension of 512.  We do not share the parameters of the transformers (the same as those in the encoder).

\section{Studying modal-incomplete datasets}
Recall that Epic-Sounds and Epic-Kitchens have \textit{modal-complete training sets}, and Ego4D-AR has \textit{modal-incomplete training set} with $r_{train} = 29\%$.
To study the effect of using MMT in the datasets with different missing modality severity levels, we create several variants of the training sets in each dataset by manually enforcing different $r_{train}>0$. We do so by randomly shuffling all modal-complete training instances and storing this order in a list. For Ego4D-AR, modal-incomplete instances ($29\%$) are placed at the very start of the list. Following this, we establish our desired missing rate $r_{train}$ by sampling from the start of this list. This method ensures that any increase in $r_{train}$ builds upon the existing set of modal-incomplete instances, meaning that if we increase $r_{train}$, we add new modal-incomplete instances to those already included, thereby maintaining a cumulative effect. We also use this protocol for the experiments in Sec.~\textcolor{red}{4.5} and Sec.~\textcolor{red}{4.6} of the main manuscript.

\begin{figure*}[t!]
\begin{center}
\includegraphics[width=\linewidth]{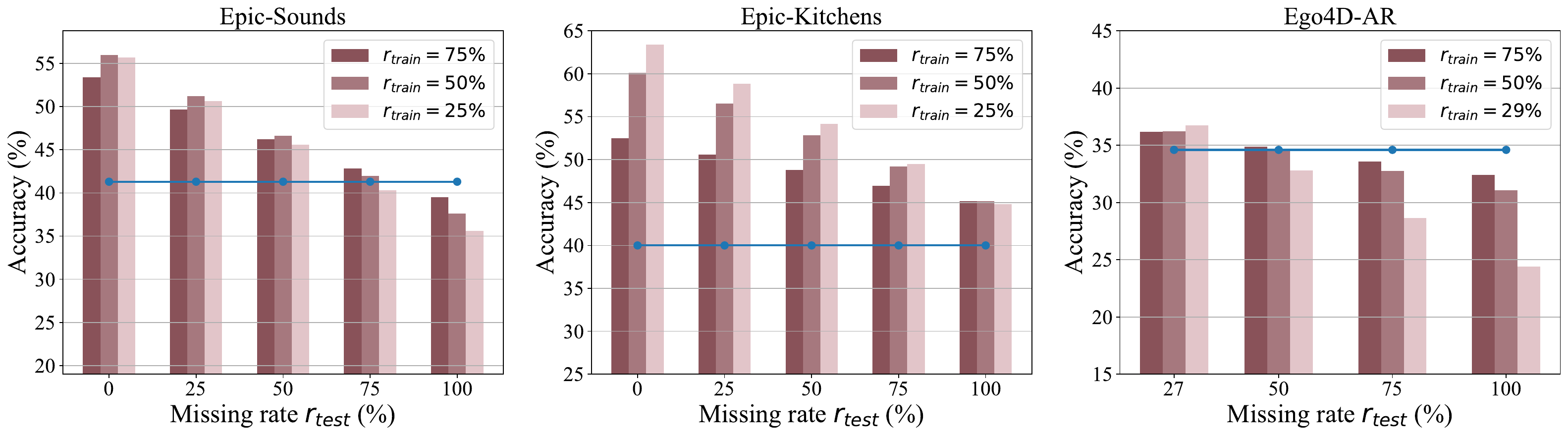}
\end{center}
\small
\caption{\textbf{Missing modality rate in training data $r_{train}$ \vs accuracy.} Our MMT effectively deals with missing modalities under several $r_{train}$ regimes. For simplicity, $p = 0$ for all experiments. 
}\vspace{-10pt}
\label{figure:ablate_r_train}
\end{figure*}

For Epic-Kitchens and Epic-Sounds we use $r_{train} \in \{25\%, 50\%, 75\%\}$, and $r_{train} \in \{29\%, 50\%, 75\%\}$ for Ego4D-AR. Figure~\ref{figure:ablate_r_train} shows the results of training our model with MMT in the datasets with different missing modality rates in the training data $r_{train}$. As can be seen, severe modal incompleteness negatively affects the performance. Nevertheless, overall, our models bring the performance closer to the unimodal for higher $r_{test}$ and still allow the model to benefit from all training samples. 

\section{Bottlenecks condensing the information.}

As mentioned in the paper, MBT uses a small set of learnable fusion tokens to exchange information between the modalities. One might wonder if the ability of the bottleneck to "condense" the information already plays some role in dealing with missing inputs. To investigate this, Table~\ref{tab:bottle_vs_self} compares training a vanilla self-attention model with $L_f = 0$ (row 1) with the training MBT with the same fusion layer (row 2). Table~\ref{tab:bottle_vs_self} shows that vanilla self-attention performs poorly in the multimodal setup with complete modalities, losing 10 points, compared to the bottleneck fusion. Surprisingly, when the modality inputs are incomplete, vanilla self-attention performs better (20.5\% vs 13.6\% with $r_{test} = 100\%$). This shows that while the bottlenecks are effective for multimodal fusion, they are sensitive to missing modalities, and they must be adapted to address incomplete modalities.
\begin{table}[]
\setlength\tabcolsep{3pt}
    \centering
    \small
\begin{tabular}{c|c|c}

\textbf{Architecture}   & \textbf{Audio $r_{test}$}    & \textbf{Accuracy} \\ \hline
\multirow{4}{*}{Self-attention}      & 0\%          & 45.3\%               \\ 
     & 50\%           &   32.7\% \\
     & 75\%           & 26.7\%    \\      & 100\%            & 20.5\%    \\
                                    \hline

\multirow{4}{*}{Bottleneck}        & 0\%       & 55.5\%       \\
  & 50\%           & 34.3\%   \\
  & 75\%           & 23.7\%   \\
  & 100\%            & 13.6\%   \\            
                                                          \midrule
Video-only  &  0\% - 100\% & 41.4\% \\                                                          
\end{tabular}
\caption{\textbf{Bottleneck fusion \vs vanilla self-attention with early fusion in Epic-Sounds dataset.} 
}
\label{tab:bottle_vs_self}
\end{table}


\newpage
\bibliographystyle{splncs04}
\bibliography{main}